%% file: acl_latex.tex
\documentclass[11pt]{article}

\usepackage[preprint]{acl}

\usepackage{times}
\usepackage{latexsym}

\usepackage[T1]{fontenc}

\usepackage[utf8]{inputenc}

\usepackage{microtype}

\usepackage{inconsolata}

\usepackage{graphicx}

\usepackage{booktabs}
\usepackage{amsmath,amssymb}
\usepackage{multirow}
\usepackage{xcolor}
\usepackage{colortbl}
\usepackage{tabularx}

\newif\ifshowrev
\showrevfalse                       %
\newcommand{\rev}[1]{\ifshowrev{\color{orange}#1}\else{#1}\fi}

\title{Seeing Is Not Sharing: Some Vision-Language Models Overestimate Common Ground in Asymmetric Dialogue}

\author{Nan Li, Albert Gatt, Massimo Poesio\\
        Utrecht University, Utrecht, The Netherlands\\
        \texttt{\{n.li, a.gatt, m.poesio\}@uu.nl}}

\begin{document}
\maketitle
\begin{abstract}
\input{sections/abstract}
\end{abstract}

\input{sections/introduction}

\input{sections/related-work}
\input{sections/experiment}
\input{sections/results}
\input{sections/discussion}
\input{sections/conclusion}

\input{sections/limitations}
\input{sections/acknowledgement}
\input{sections/availibility}

\bibliography{custom}

\input{sections/appendix}

\end{document}

%% file: sections/abstract.tex
In collaborative dialogue, shared perception does not guarantee shared interpretation. Mutual understanding must be established through interaction.
We investigate whether vision-language models (VLMs) can distinguish what \emph{could} be shared from what \emph{has been} shared between dialogue participants through grounding.
We formulate this as an interpretation-matching task on 13,077 annotated reference expressions from HCRC MapTask dialogues, and evaluate VLMs under systematically controlled manipulations of dialogue context and map-information access.
Our results show that providing authentic map images improves overall performance but shifts models toward over-predicting alignment. Textual descriptions of the same map content reproduce this bias, while non-informative images suppress alignment predictions entirely, indicating that the bias is driven by task-relevant map content, not the visual channel. This improvement comes at the cost of degraded accuracy on non-aligned cases.
Calibration analysis and reference-chain tracking further suggest that models rely on static referential cues on the maps rather than tracking how grounding unfolds through dialogue history.
\rev{We observe these patterns most clearly in Qwen3-VL-8B-Instruct and, to varying degrees, in four additional models from two architecture families.}
\rev{In models that exhibit the bias, map content, whether presented visually or textually, is treated as evidence of mutual understanding, conflating potential with established common ground.}

%% file: sections/introduction.tex
\begin{figure*}[t]
    \centering
    \includegraphics[width=\textwidth]{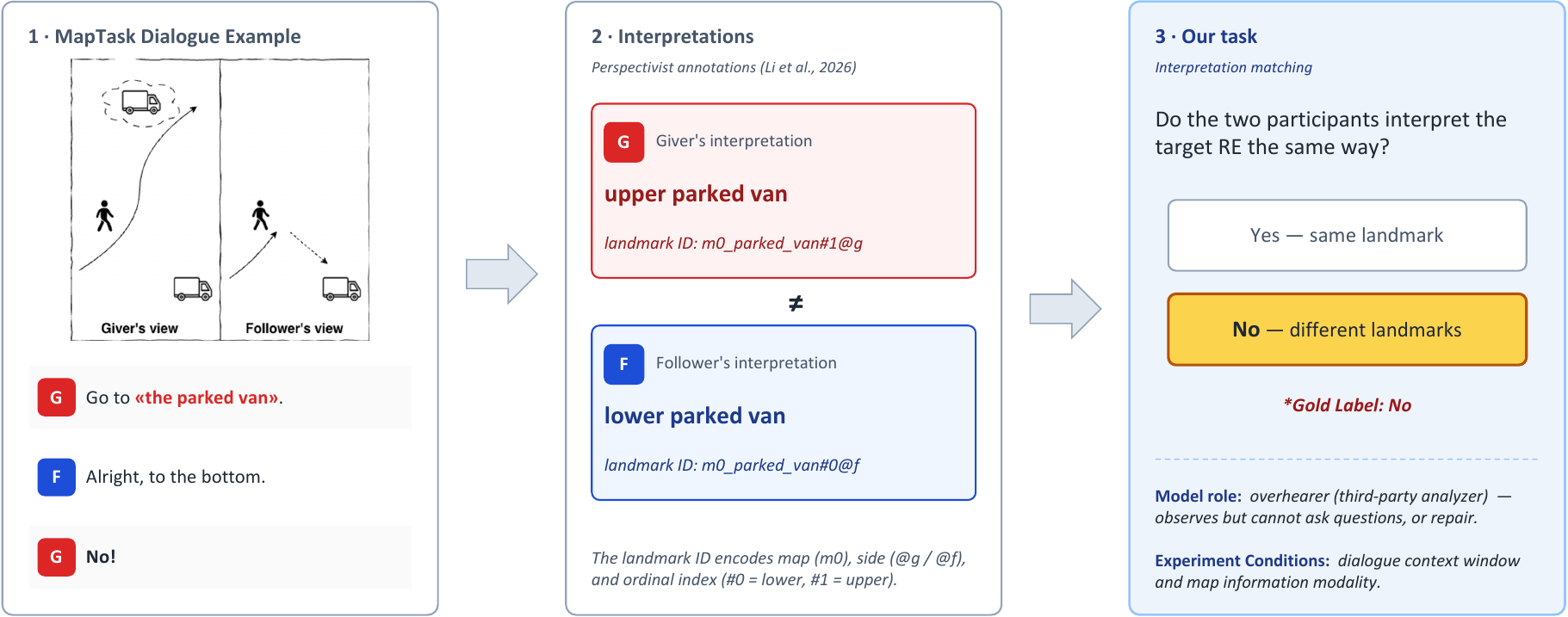}
    \caption{The interpretation matching task, illustrated on a misalignment example. \textbf{Panel 1} shows a simplified example of a MapTask dialogue with the target RE \emph{the parked van}: the giver's map contains two parked vans while the follower's map has only one. \textbf{Panel 2} shows the perspectivist annotations~\citep{li2025grounded}: the giver grounds \emph{the parked van} to the upper van (\texttt{m0\_parked\_van\#1@g}, see Appendix~\ref{sec:app-dataset} for the definition about the landmark ID) and the follower to the lower one (\texttt{m0\_parked\_van\#0@f}). \textbf{Panel 3} formulates our task: given the map(s) and dialogue, we ask the model to decide whether the two participants' interpretations of the marked RE match. The gold label here is \textsc{No}.}
    \label{fig:task-example}
\end{figure*}

\section{Introduction}\label{sec:intro}

In everyday collaborative situated dialogue, two people attending to the situation may extract different information from it or interpret the same information in different ways.
Grounding, the incremental process by which dialogue participants establish mutual understanding, is what bridges this gap~\cite{clark1986ReferringCollaborativeProcess, clark1989ContributingDiscourse, clark1991GroundingCommunication}.
A speaker's reference expression (RE) becomes part of the common ground only after the addressee recognises, accommodates, and confirms it; until then, the RE remains \emph{potential} rather than \emph{established} shared knowledge.

\rev{While information asymmetry is present to some degree in all dialogue, in certain collaborative tasks it is introduced by design, making asymmetry a \emph{structural and controlled} feature of the task. In what follows, we use \emph{asymmetric dialogue} to refer specifically to settings where participants hold different private task-relevant information by design, and where neither can directly access the other's.}
The HCRC MapTask~\cite{anderson1991hcrc} is a canonical instance: two participants navigate a route using maps that differ in landmark placement, landmark names, or the number of identically-named landmarks, so that apparent agreement can mask genuine divergence in interpretation.
\rev{Because the discrepancies are designed into the maps, landmarks visible on both maps create \emph{potential} referential overlap whose resolution depends entirely on the grounding process in the dialogue. A model evaluating such dialogue from the outside, as an overhearer with access to one or both maps, faces the same asymmetry: it can observe what \emph{could} be shared, but must infer from the dialogue what \emph{has been} shared.}
Recent perspectivist annotation work \cite{li2025grounded} has made it possible to evaluate, for each RE, whether two MapTask participants actually share the same grounded interpretation at a given point in the dialogue --- a judgment we call \emph{interpretation matching}.

We use this resource to investigate three research questions:
\begin{enumerate}
    \item[\textbf{RQ1}] Can large vision-language models (VLMs) capture personal interpretations of interlocutors toward the same reference expression in asymmetric dialogue?
    \item[\textbf{RQ2}] Which modality of information contributes more to VLMs' assessment of interpretation alignment?
    \item[\textbf{RQ3}] Do VLMs exhibit systematically different behaviours on different types of alignment and misalignment cases?
\end{enumerate}

We frame the task as a binary judgment --- do the two participants' interpretations of a marked RE match? --- and evaluate \rev{VLMs} under systematic manipulations of two independent variables: the amount of available dialogue context and the type of map information provided. \rev{We evaluate models from two open-source VLM families, Qwen3-VL and Gemma3, at scales from 2B to 12B parameters. Qwen3-VL-8B-Instruct, the best-performing model in preliminary evaluation, serves as the primary model for detailed condition-grid analysis; the remaining models are compared on a baseline grid to assess cross-model generality (\S\ref{sec:exp-models}, \S\ref{sec:model-comparison}).}

Our main findings are:
\begin{itemize}
    \item Providing authentic map images shifts the model toward over-predicting alignment.
    The model treats landmark co-presence as evidence of mutual understanding, conflating \emph{what could be shared} with \emph{what has been shared}.
    Textual descriptions of the same map content reproduce this bias, showing that it is driven by task-relevant map content rather than by the visual channel.
    \item Non-informative visual inputs (blank maps, shuffled landmarks) do not reproduce the bias; they make the model \emph{more conservative}, not more prone to predicting alignment. The bias therefore possibly requires map \emph{content}, whether delivered visually or textually.
    \item Calibration and reference-chain analysis converge on the same explanation: the model relies more on static referential cues on the maps rather than tracking how grounding unfolds through dialogue history.
\end{itemize}

\rev{Our work contributes (1) an evaluation methodology for probing VLMs' ability to distinguish potential from established common ground in information-asymmetric dialogue; and (2) an empirical characterisation of a model-dependent failure mode, observed most clearly in Qwen3-VL-8B, in which VLMs conflate possible referential overlap with communicative alignment.}

%% file: sections/related-work.tex
\section{Related Work}\label{sec:related-work}

\paragraph{Grounding and Overhearers}
Common ground is built incrementally through interaction: interlocutors negotiate, confirm, and repair meaning, rather than receiving it directly from shared perceptual access~\cite{clark1986ReferringCollaborativeProcess,clark1991GroundingCommunication}.
A well-established consequence is the \emph{overhearer illusion}: overhearers can hear every word but, lacking the ability to contribute grounding acts, reach systematically weaker interpretations than addressees~\cite{schober1989UnderstandingAddresseesOverhearers}.
This asymmetric performance carries over to language models: dialogue systems trained and evaluated on static transcripts are structurally overhearers, which shapes what they can learn about grounding and clarification~\cite{madureira-schlangen-2024-couldnt}.
Our evaluation makes that stance explicit, placing VLMs as overhearers of asymmetric human dialogue and testing whether they mistake co-presence of potential referents on the maps, delivered either visually or as text, for established shared interpretation.

\paragraph{Reference in Dialogue and VLM Evaluation}
Reference corpora have long served as testbeds for how interlocutors build shared interpretations through repeated mention, partial information, or clarification~\cite{anderson1991hcrc, haber2019photobook,udagawa2019NaturalLanguageCorpus, chiyah-garcia-etal-2023-referring}, yet each RE is typically annotated with a single gold referent, implicitly assuming speaker-addressee convergence.
Under information asymmetry this breaks: the perspectivist annotation of \citet{li2025grounded} shows that the two participants can hold distinct grounded interpretations even under apparent agreement, the judgement of which overhearers make hard.
Recent VLM evaluations report systematic gaps with humans as well: VLM overhearers underperform human matchers on referential dialogue and do not improve with repeated discussion~\cite{wang-etal-2025-lvlms}, and VLM participants fail to entrain, form conceptual pacts, or initiate grounding acts as humans~\cite{zeng2026lvlms, shaikh-etal-2025-navigating}.
We directly ask whether a VLM, given the same asymmetric evidence as the participants, can recognise that they have not yet reached a shared interpretation, evaluating VLMs on perspectivist MapTask annotations under a variety of controlled map-information conditions.

%% file: sections/experiment.tex
\section{Experimental Setup}\label{sec:exp}

\subsection{Dataset}\label{sec:exp-dataset}

We use a published corpus of perspectivist annotations of the HCRC MapTask corpus \citep{li2025grounded} that separately records, for each RE, the speaker's intended landmark and the addressee's interpreted landmark.
The dataset comprises 13,077 annotated REs from 128 HCRC MapTask dialogues. Each dialogue involves two participants (a giver and a follower) collaborating to reproduce a route on the follower's map under the giver's guidance, with slightly different maps. The discrepancies can include landmark name differences, missing landmarks, and differences in quantity, creating a rich environment for misalignment in grounding. See Appendix~\ref{sec:app-dataset} for more dataset details.

\subsection{Task Design}\label{sec:exp-task}

We want to evaluate VLMs' \emph{interpretation matching} ability, and formulate this as a binary judgment task:
\emph{Given a MapTask dialogue excerpt containing one marked RE, the model must decide whether the two participants currently share the same grounded interpretation of that expression or not.}
Figure~\ref{fig:task-example} illustrates a misalignment case, where the giver and follower ground \emph{the parked van} to different landmarks.

An instance is labelled \textsc{Yes} (aligned) when the two participants' landmark IDs match; and \textsc{No} (not aligned) otherwise, covering both pending states (not yet grounded) and misunderstandings (grounded to different landmarks).
The ground truth class distribution of the corpus is imbalanced: 72.1\% aligned (\textsc{Yes}) and 27.9\% not aligned (\textsc{No}).

We manipulate two independent variables to investigate how information access shapes model judgments: \emph{text access} (the dialogue context window) and \emph{map access} (the map-information modality).

\paragraph{Text access (dialogue context window)}
We vary how much dialogue context the model receives via four windows of increasing size: \textbf{curL} (current transaction, up to and including the line containing the target RE), \textbf{curT} (current transaction in full), \textbf{startL} (dialogue from beginning through the line containing the target RE), and \textbf{startT} (dialogue from beginning through the end of the current transaction).
A \emph{transaction} is a human-annotated MapTask dialogue excerpt that typically corresponds to a sequence of movements from one landmark to another.

These windows allow us to test whether (1) broader dialogue history, which encodes prior grounding episodes, and (2) future interactions, which usually encode repair sequences and clarification exchanges, help the model track grounding.

\paragraph{Map access (map-information modality)}
We vary what map information the model receives.
In the \emph{baseline condition grid}, conditions include: \textbf{Text-only} (no map information), \textbf{Both maps} (authentic giver and follower map images), and \textbf{Giver-only / Follower-only} (a single authentic map image).

To further investigate the impact of map information and disentangle content from input channel, the \emph{full-grid} modality experiment adds four conditions. Two are textual: \textbf{Text-landmark-names} (a textual list of landmark names on each map) and \textbf{Text-discrepancy-detail} (textual descriptions about the map discrepancies). Two are non-informative visual controls: \textbf{Blank maps} (uniform empty images, 1024$\times$1024, RGB 128/128/128) and \textbf{Shuffled maps} (map images with landmarks from an unrelated map pair). Appendix~\ref{sec:app-textual-mapinfo} shows concrete example fillings of \texttt{\$\{map\_access\}} for the two textual conditions.

\begin{table*}[t]
\centering
\small
\begin{tabular}{ll cccc c}
\toprule
\textbf{Map access} & \textbf{Text} & \textbf{Accuracy} & \textbf{F1$_{\text{macro}}$} & \textbf{Recall$_{\text{pos}}$} & \textbf{Recall$_{\text{neg}}$} & \textbf{Yes-rate} \\
\midrule
\multirow{4}{*}{Text-only}
  & curL   & .442 & .439 & .261 & .910 & .213 \\
  & curT   & .639 & .604 & .647 & .616 & .574 \\
  & startL & .533 & .532 & .409 & .855 & .336 \\
  & startT & .614 & .591 & .590 & .677 & .515 \\
\midrule
\multirow{4}{*}{Both maps}
  & curL   & .627 & .593 & .633 & .609 & .566 \\
  & curT   & .654 & .618 & .665 & .625 & .584 \\
  & startL & .694 & .630 & .769 & .501 & .694 \\
  & startT & .737 & .671 & .822 & .518 & .727 \\
\midrule
\multirow{4}{*}{Giver-only}
  & curL   & .626 & .586 & .650 & .565 & .590 \\
  & curT   & .688 & .632 & .747 & .536 & .668 \\
  & startL & .723 & .644 & .827 & .454 & .749 \\
  & startT & .756 & .669 & .879 & .436 & .791 \\
\midrule
\multirow{4}{*}{Follower-only}
  & curL   & .619 & .569 & .663 & .504 & .617 \\
  & curT   & .653 & .594 & .716 & .490 & .658 \\
  & startL & .718 & .632 & .832 & .422 & .761 \\
  & startT & .743 & .650 & .872 & .408 & .794 \\
\bottomrule
\end{tabular}
\caption{Baseline results across text-access and map-access conditions for the model Qwen3-VL-8B-Instruct. Best F1$_{\text{macro}}$ per map-access level is \textbf{startT} in all map conditions; for text-only, curT slightly outperforms startT.}
\label{tab:baseline}
\end{table*}

The blank-map and shuffled-map conditions serve as controls: if the over-alignment bias is a generic multimodal artifact, these conditions should reproduce it; if it is content-driven, they should not.

\paragraph{Prompt design}
We apply zero-shot learning. The model receives a system prompt framing it as a dialogue analysis expert overhearing a MapTask dialogue. The prompt further describes map asymmetry (landmarks may be missing, duplicated, or placed differently), and defines the interpretation-matching task.
The user prompt provides the target RE, the dialogue context with the target RE wrapped in \texttt{<<...>>} markers, and (where applicable) map images.
Full prompt templates are given in Appendix~\ref{sec:app-prompt}.

\subsection{Models and Inference}\label{sec:exp-models}

\rev{We select models from two open-source VLM families that represented the state of the art at the time of our experiments: Qwen3-VL~\cite{bai2025qwen3vltechnicalreport} and Gemma3~\cite{kamath2025gemma}. Within each family, we evaluate models ranging from 2B to 12B parameters (Qwen3-VL-2B/4B/8B-Instruct; Gemma3-4B/12B-it), constrained by the memory of a single A100 GPU. All models are instruction-tuned, non-thinking versions.}
\rev{In preliminary evaluation across all five models, Qwen3-VL-8B-Instruct achieved the highest macro-F1, so we use it as the primary model for the full condition-grid analysis. The remaining four models are evaluated on the baseline condition grid to assess generality (\S\ref{sec:model-comparison}).}

We use vLLM~\cite{kwon2023efficient} to deploy the models and to accelerate the inference progress. We use greedy decoding (temperature 0) with constrained output via vLLM logit masking to valid \textsc{Yes}/\textsc{No} tokens, ensuring valid binary responses. We set \texttt{random\_seed} to 42 in vLLM's sampling parameters to ensure reproducibility.

Map images are resized to a maximum side length of 1024 pixels and delivered as PNG. We selected 1024 pixels as the minimum resolution at which Qwen3-VL-8B-Instruct reliably identified landmark names, icons, and locations in preliminary checks.

\subsection{Evaluation}\label{sec:exp-eval}

We report accuracy, macro-averaged F1, per-class recall (recall$_{\text{pos}}$ for aligned, recall$_{\text{neg}}$ for not-aligned), and the model's \emph{yes-rate} (proportion of \textsc{Yes} predictions) as a measure of response bias.
In addition, we conduct three further analyses in \S\ref{sec:discussion}: calibration analysis using token-level logits (\S\ref{sec:calibration}), a status-level breakdown that decomposes performance by grounding state (\S\ref{sec:status-tradeoff}), and reference-chain tracking that examines how predictions evolve across repeated mentions of the same landmark (\S\ref{sec:ref-chains}).

%% file: sections/results.tex
\section{Results}\label{sec:results}

Table~\ref{tab:baseline} presents the main results across the baseline conditions. \rev{All results in this section use Qwen3-VL-8B-Instruct; cross-model generality is assessed in \S\ref{sec:model-comparison}.} Here are the key findings:

\paragraph{F1: Map access improves detection but shifts the model toward over-predicting alignment.}
Comparing within the same model (Qwen3-VL-8B), at startT, adding both maps improves F1$_{\text{macro}}$ from .591 (text-only) to .671 (both maps), a gain of .080.
However, this improvement is driven by a dramatic shift in recall profile: recall$_{\text{pos}}$ rises from .590 to .822, while recall$_{\text{neg}}$ drops from .677 to .518.
\rev{The yes-rate shifts from .515 to .727, exceeding the gold base rate of .721. The evidence for over-prediction is not the yes-rate alone (which is close to the base rate) but the \emph{direction of the recall trade-off}: map access systematically pushes recall$_{\text{neg}}$ down while inflating recall$_{\text{pos}}$, meaning the model sacrifices its ability to detect non-aligned cases in favour of aligned ones. Calibration analysis in \S\ref{sec:calibration} and the status-level breakdown in \S\ref{sec:status-tradeoff} further confirm this: map conditions produce confident errors specifically on gold-\textsc{No} instances.}
We analyse the status-level consequences of this shift in \S\ref{sec:status-tradeoff}, including its differential impact on aligned, pending, and misunderstood instances.

Single-map conditions amplify the bias.
The giver-only condition at startT achieves similar F1$_{\text{macro}}$ (.669) to both maps (.671), but with an even higher yes-rate (.791) and lower recall$_{\text{neg}}$ (.436).
The follower-only condition shows a comparable pattern (yes-rate .794, recall$_{\text{neg}}$ .408).
Access to either map is sufficient to trigger the over-alignment bias, and providing both maps actually moderates it slightly by introducing cross-map discrepancy evidence.

\begin{table}[htbp]
\centering
\small
\begin{tabular}{l cccc}
\toprule
\textbf{Map info} & \textbf{F1$_{\text{macro}}$} & \textbf{R$_{\text{pos}}$} & \textbf{R$_{\text{neg}}$} & \textbf{Yes-rate} \\
\midrule
\multicolumn{5}{l}{\emph{Baseline}} \\
Text-only           & .591 & .590 & .677 & .515 \\
Both maps    & .671 & .822 & .518 & .727 \\
\midrule
\multicolumn{5}{l}{\emph{Textual map information}} \\
Landmark names      & .636 & .756 & .533 & .675 \\
Discrepancy Desc.  & .668 & .810 & .528 & .716 \\
\midrule
\multicolumn{5}{l}{\emph{Fake visual controls}} \\
Blank maps          & .407 & .220 & .910 & .184 \\
Shuffled maps       & .402 & .222 & .884 & .193 \\
\bottomrule
\end{tabular}
\caption{Map-information modality comparison at startT (both-maps access level). All conditions use Qwen3-VL-8B-Instruct. R$_{\text{pos}}$ = recall on aligned; R$_{\text{neg}}$ = recall on not-aligned.}
\label{tab:modality}
\end{table}

\paragraph{F2: Textual map descriptions reproduce the over-alignment bias; only content-free visual controls avoid it.}
To determine whether the over-alignment bias is driven by map \emph{content} or by the visual input channel, we compare authentic maps against textual map descriptions and non-informative visual controls.
All conditions use the same model (Qwen3-VL-8B-Instruct) at the startT text window, ensuring a clean within-architecture comparison. Table~\ref{tab:modality} shows the results.

Blank maps (yes-rate .184) and shuffled maps (yes-rate .193) produce \emph{lower} yes-rates than the text-only baseline (.515), let alone authentic maps (.727).
The model becomes more conservative when it receives images from which it cannot extract task-relevant content.
This rules out the hypothesis that the over-alignment bias is caused by the mere presence of images rather than by the information they carry.

Both textual conditions produce yes-rates (.675--.716) close to real maps (.727) and well above the text-only baseline (.515).
Macro F1$_{\text{macro}}$ for the textual conditions (.636--.668) sits just below real maps (.671) and well above text-only (.591).
The recall profiles echo real maps: elevated recall$_{\text{pos}}$ (.756--.810) and reduced recall$_{\text{neg}}$ (.528--.533), the same directional shift as real maps (.822 / .518), only slightly less extreme.

The same landmark information --- which landmarks appear on which maps and where they differ --- triggers over-alignment whether it is presented visually or textually.
The over-alignment bias is therefore about \emph{what the model learns about the scene}, not about the visual channel per se.
The visual presentation does contribute a small additional shift (a few points on yes-rate and on recall$_{\text{pos}}$) relative to the textual presentation of the same content, consistent with spatial co-presence in images being a slightly stronger perceptual cue, but this residual effect is small compared to the content effect itself.

The conditions thus split into two groups by yes-rate:
inputs with task-relevant map content (real maps .727, textual descriptions .675--.716) over-predict alignment;
inputs without map content (text-only .515, blank maps .184, shuffled maps .193) do not.
Map content thus seems to drive the bias, while the visual presentation channel has only a secondary amplifying effect.

\paragraph{F3: Broader dialogue context helps, but this is mitigated by map access.}
The text-window effect (curL $\to$ startT) produces a .152 F1$_{\text{macro}}$ gain in the VL text-only condition, but only .078 under both maps.
This means when the model has no access to maps, giving it more dialogue context helps a lot more. When it already has both maps, extra dialogue context still helps, but much less.
This smaller gain suggests that, under map access, the model relies more heavily on static referential cues from the maps and benefits less from additional dialogue evidence about how grounding unfolds over time.

Future dialogue interactions also help. Across all map conditions, both curT $\to$ startT and curL $\to$ startL produce consistent gains in F1$_{\text{macro}}$.
This means that subsequent turns often contain useful repair, confirmation, or clarification evidence that makes the grounding outcome more legible.

%% file: sections/discussion.tex
\section{Further Analysis and Discussion}\label{sec:discussion}

The preceding results show that map access improves overall performance while amplifying the over-alignment bias, and that this bias is driven by task-relevant map content rather than by the visual channel itself.
We now probe the mechanism behind this pattern through calibration analysis (\S\ref{sec:calibration}), a status-level trade-off analysis (\S\ref{sec:status-tradeoff}), reference-chain tracking (\S\ref{sec:ref-chains}), and cross-model comparison (\S\ref{sec:model-comparison}).

\subsection{Over-Prediction and Calibration}\label{sec:calibration}

Since we use vLLM's constrained decoding to force a single \textsc{Yes}/\textsc{No} token, the cumulative log-probability of the generated token directly gives the model's \emph{confidence}: $\text{conf} := \exp(\text{logprob})$.
We also compute Expected Calibration Error (ECE)~\cite{naeini2015obtaining,guo2017calibration} to measure how calibration quality varies depending on whether the model's default response happens to be correct.

\begin{table}[h]
\centering
\small
\begin{tabular}{l cccc}
\toprule
\textbf{Condition} & \textbf{ECE} & \textbf{ECE$_\text{yes}$} & \textbf{ECE$_\text{no}$} & \textbf{Conf.} \\
\midrule
Text-only          & .249 & .263 & .235 & .863 \\
Both maps          & .174 & .094 & .403 & .912 \\
Giver-only         & .172 & .057 & .484 & .927 \\
Follower-only      & .185 & .061 & .524 & .929 \\
\bottomrule
\end{tabular}
\caption{Calibration by gold label at startT ($n$ = 13,077). All conditions use Qwen3-VL-8B-Instruct. ECE$_\text{yes}$/ECE$_\text{no}$ = ECE on gold-\textsc{Yes}/gold-\textsc{No} instances. Conf.\ = mean prediction confidence.}
\label{tab:calibration}
\end{table}

Table~\ref{tab:calibration} shows the results with Qwen3-VL-8B-Instruct at startT across all 13,077 instances. Here are the key findings:

\paragraph{Map conditions are better calibrated on aligned instances but miscalibrated on non-aligned ones.}
Under both-maps, the model is biased toward \textsc{Yes} (yes-rate .727) and is well-calibrated on gold-\textsc{Yes} instances (ECE$_\text{yes}$ = .094) but badly miscalibrated on gold-\textsc{No} instances (ECE$_\text{no}$ = .403).
The text-only condition, which is not strongly biased toward either class (yes-rate .515), is moderately calibrated on both classes (ECE$_\text{yes}$ = .263, ECE$_\text{no}$ = .235).
The asymmetry is sharpest under single-map conditions: follower-only achieves ECE$_\text{yes}$ = .061 but ECE$_\text{no}$ = .524.

\paragraph{Maps make the model more confident, and more confidently wrong on non-aligned instances.}
Mean confidence rises modestly with map access (.863 $\to$ .912 / .927 / .929), but the distribution of that confidence shifts: the ECE gap between gold-\textsc{Yes} and gold-\textsc{No} instances widens from .028 (text-only) to .463 (follower-only). On gold-\textsc{Yes} instances the model is confident and right; on gold-\textsc{No} instances it is equally confident but wrong. In other words, when the model gets non-aligned instances wrong, it is confidently wrong. Map evidence drives the model to predict \textsc{Yes} with high certainty even when alignment does not hold.
\rev{This class-conditioned miscalibration is the strongest evidence for over-prediction: on gold-\textsc{No} instances, the model is not merely wrong but \emph{confidently} wrong, indicating that map content drives spurious certainty rather than merely shifting a threshold.}

\subsection{The Status-Level Trade-Off}\label{sec:status-tradeoff}

The overall improvement from map access conceals an asymmetric trade-off.
Table~\ref{tab:status} decomposes accuracy by the gold grounding status of each RE: \emph{aligned} (gold \textsc{Yes}; $n$ = 9,435), \emph{pending} (not yet grounded; gold \textsc{No}; $n$ = 3,403), or \emph{misunderstood} (grounded to different landmarks; gold \textsc{No}; $n$ = 239).

\begin{table}[h]
\centering
\small
\begin{tabular}{l ccc}
\toprule
\textbf{Condition} & \textbf{Aligned} & \textbf{Pending} & \textbf{Misund.} \\
\midrule
\multicolumn{4}{l}{\emph{Baseline}} \\
Text-only       & .590 & .691 & .473 \\
Both maps       & .822 & .523 & .456 \\
Giver-only      & .879 & .441 & .372 \\
Follower-only   & .872 & .419 & .255 \\
\midrule
\multicolumn{4}{l}{\emph{Textual map information}} \\
Landmark names  & .756 & .544 & .372 \\
Disc.\ detail   & .810 & .540 & .368 \\
\midrule
\multicolumn{4}{l}{\emph{Fake visual controls}} \\
Blank maps      & .220 & .913 & .866 \\
Shuffled maps   & .222 & .886 & .858 \\
\bottomrule
\end{tabular}
\caption{Accuracy by gold grounding status at startT (both-maps access level). Each cell shows the fraction of correct predictions within that status group. All conditions use Qwen3-VL-8B.}
\label{tab:status}
\end{table}

\paragraph{Maps boost aligned accuracy at the cost of pending and misunderstood accuracy.}
Maps boost aligned accuracy by 23--29 percentage points (.590 $\to$ .822 / .872 / .879) while simultaneously dropping pending accuracy by 17--27 points (.691 $\to$ .419 / .441 / .523; McNemar $p < 10^{-6}$ for all map conditions).
On misunderstood REs, the effect is condition-dependent: the both-maps drop (.473 $\to$ .456) is not significant ($p = 0.724$, $n = 239$), but single-map conditions show large significant declines. Follower-only drops to .255 ($p = 3 \times 10^{-9}$) and giver-only to .372 ($p = 8 \times 10^{-3}$).

Both maps moderate the misunderstood collapse.
Having both maps is actually the \emph{least extreme} map condition on misunderstood (0.456 vs.\ 0.372 giver-only vs.\ 0.255 follower-only), likely because cross-map discrepancies provide corrective evidence that moderates the bias relative to single-map conditions.
The accuracy drop on pending cases, by contrast, is broad and significant for \emph{all} map conditions.

\paragraph{Textual descriptions follow the same trade-off as authentic maps.}
Textual map descriptions show the same directional profile as authentic maps relative to the text-only baseline (.590 / .691 / .473 for aligned / pending / misunderstood): aligned accuracy rises sharply while pending and misunderstood accuracy drop.
Discrepancy-detail reaches .810 / .540 / .368 and landmark names .756 / .544 / .372 --- close to both-maps (.822 / .523 / .456) on aligned and pending, but notably \emph{below} both-maps on misunderstood.
Their F1$_{\text{macro}}$ (.636 / .668) sits just under both-maps (.671).

Fake visual controls, by contrast, push the profile in the opposite direction: blank and shuffled maps produce near-identical hyper-conservative patterns (.220 / .222 aligned, .913 / .886 pending, .866 / .858 misunderstood), collapsing into a near-constant \textsc{No} response (yes-rates .184 / .193).

The conditions therefore form two groups: content-rich inputs (real maps and textual descriptions) over-predict alignment, while content-free visual inputs (blank, shuffled) amplify caution to the opposite extreme.
Telling the model what is on the maps or showing it produces similar behaviour; it is the presence of task-relevant content about potential common ground that drives the trade-off, not the modality.

\paragraph{The model confuses potential with established common ground.}
Map content tells the model what \emph{could} be shared (landmarks appearing on both maps); dialogue history tells it what \emph{has been} shared (interpretations established through interaction).
The model over-weights the former.
This is the computational analogue of the \emph{overhearer's illusion}: overhearers systematically overestimate their understanding of a conversation because they have access to referential context but lack the interactive grounding process that establishes mutual understanding between participants.
The model is structurally an overhearer, which means it observes what both participants could share but cannot well assess what they have confirmed through dialogue.

\subsection{Reference-Chain Analysis}\label{sec:ref-chains}

\begin{figure}[h]
\centering
\includegraphics[width=\columnwidth]{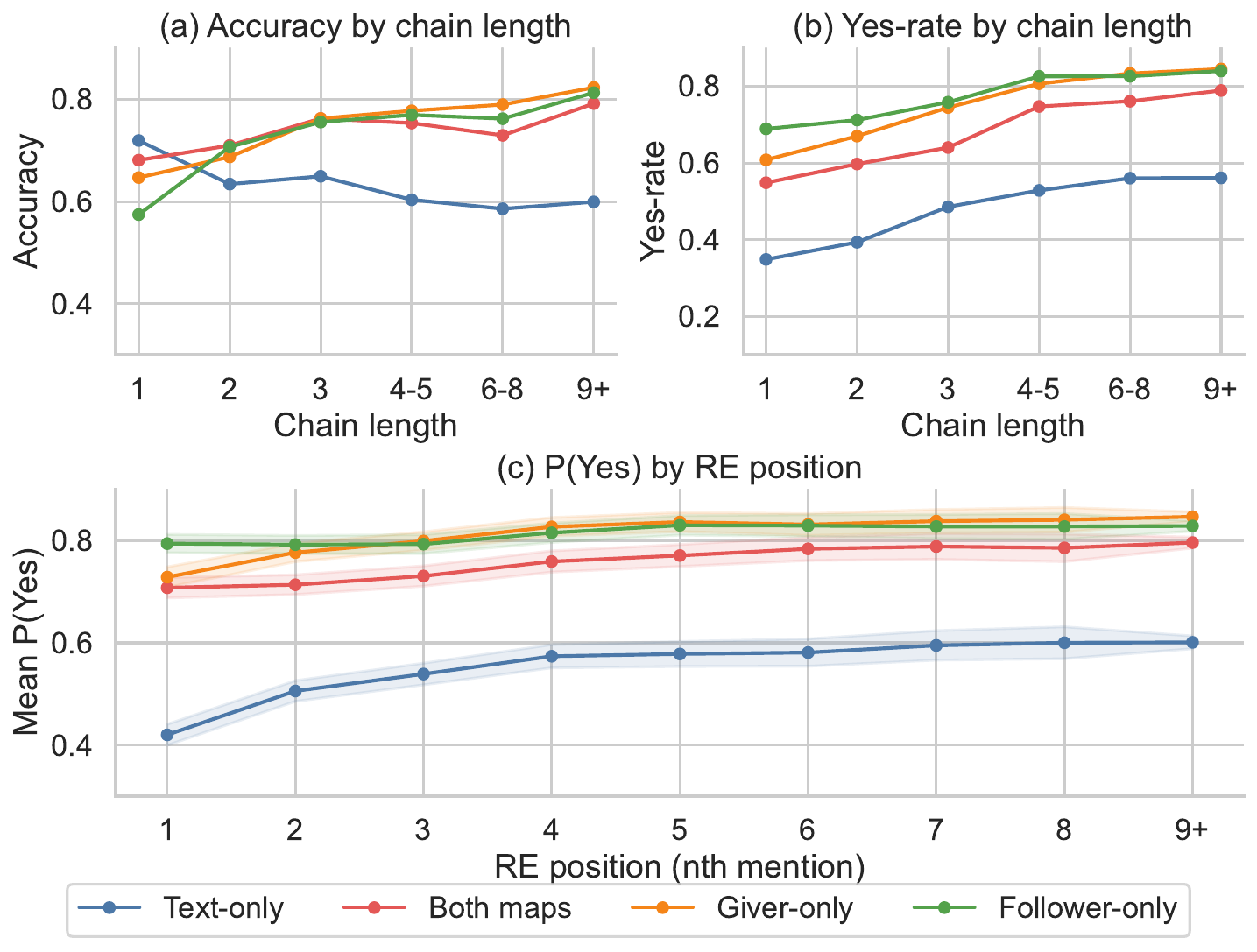}
\caption{(a) Accuracy and (b) yes-rate by reference-chain length; (c) mean $P(\textsc{YES})$ by RE position (nth mention within a chain), with 95\% CI shading. Map conditions maintain high accuracy but with steadily increasing yes-rate, and $P(\textsc{YES})$ rises with RE position in most situations.}
\label{fig:chain-length}
\end{figure}

We use the \emph{reference chains} defined in \citet{li2025grounded}, sequences of mentions of the same landmark within a dialogue, to test whether repeated mention helps the model converge on the correct judgment.
We group 1,665 chains into six buckets  by \emph{chain length}, the number of mentions of that landmark in the dialogue (1, 2, 3, 4--5, 6--8, 9+), and report accuracy together with yes-rate, because chain-length effects are partly driven by response bias that macro F1 alone obscures. See Figure~\ref{fig:chain-length}.

\paragraph{Over-prediction of alignment grows with repeated mention.}
Map conditions maintain high accuracy across chain lengths (both-maps: .681 at length 1 $\to$ .791 at 9+), while text-only degrades (.719 $\to$ .599; Figure~\ref{fig:chain-length}a).
However, this comes with steadily rising yes-rates under map conditions (both-maps: .549 $\to$ .789; single-map reaches .840--.845; Figure~\ref{fig:chain-length}b).
Since aligned REs dominate at longer chain lengths (post-grounding mentions accumulate), this inflation lets maps appear more accurate than they are.
The same pattern emerges within chains: $P(\textsc{YES})$ rises steadily with RE position in most conditions, especially at the early mentions (Figure~\ref{fig:chain-length}c) --- text-only from .420 at the first mention to .601 at 9+, and map conditions (.708--.794) by +.034 to +.118.

\subsection{Cross-Model Comparison}\label{sec:model-comparison}

To assess whether the above findings and model behavioural patterns generalise beyond a single model, we evaluate four additional VLMs from two VL model families on the baseline condition grid at startT. Table~\ref{tab:models} shows the results.

\begin{table}[h]
\centering
\small
\setlength{\tabcolsep}{3pt}
\begin{tabular}{l cc cccc}
\toprule
 & \multicolumn{2}{c}{\textbf{Text-only}} & \multicolumn{4}{c}{\textbf{Both maps}} \\
\cmidrule(lr){2-3} \cmidrule(lr){4-7}
\textbf{Model} & F1$_\text{m}$ & YR & F1$_\text{m}$ & YR & ECE & Conf \\
\midrule
Qwen3-VL-2B  & .561 & .483 & .566 & .707 & .144 & .789 \\
Qwen3-VL-4B  & .518 & .368 & .411 & .225 & .494 & .907 \\
Qwen3-VL-8B  & .591 & .515 & .671 & .727 & .174 & .912 \\
\midrule
Gemma-3-4B   & .445 & .279 & .510 & .400 & .457 & .973 \\
Gemma-3-12B  & .322 & .107 & .416 & .231 & .531 & .948 \\
\bottomrule
\end{tabular}
\caption{Cross-model comparison at startT on the full dataset ($n$ = 13,077). F1$_\text{m}$ = F1$_{\text{macro}}$; YR = yes-rate.}
\label{tab:models}
\end{table}

\paragraph{Models respond to map information in qualitatively different ways.}
Adding maps increases the yes-rate for Qwen3-VL-8B (.515 $\to$ .727), Qwen3-VL-2B (.483 $\to$ .707), and Gemma-3-4B (.279 $\to$ .400), the same over-alignment direction observed in the primary experiments.
However, Qwen3-VL-4B responds in the \emph{opposite} direction: maps make it more conservative (yes-rate .368 $\to$ .225, F1$_{\text{macro}}$ .518 $\to$ .411).
Gemma-3-12B is extremely conservative across all conditions (yes-rate .107 text-only, .231 both-maps), barely engaging with the task.
These divergent responses suggest that the over-alignment bias is not a universal property of vision-language architectures but depends on model-specific factors.

\rev{One likely contributor to Gemma3's weaker performance is its limited ability to parse the information-dense hand-drawn MapTask map images. The two families differ substantially in how they encode visual input. Qwen3-VL uses a native dynamic-resolution vision encoder~\cite{bai2025qwen3vltechnicalreport}: images are tiled into patches at their input resolution, producing a variable number of visual tokens that scales with image area. For our 791$\times$1024 map images, this yields hundreds of tokens with 2D-RoPE-based spatial positional encoding, preserving fine-grained detail such as small landmark labels and icons.
Gemma3 uses a SigLIP-based vision encoder~\cite{kamath2025gemma} that resizes images to a fixed 896$\times$896 resolution and average-pools the patch representations down to a budget of 256 tokens per image, regardless of image size or content complexity. This aggressive compression likely discards spatial detail that is critical for reading the MapTask maps.
Prior to the main experiments, we ran a sanity check in which all five models were prompted to list landmark names and describe their spatial positions on each of the 32 map images (Appendix~\ref{sec:app-map-reading}). Qwen3-VL models achieved 88--90\% F1 on landmark identification, compared to 81--82\% for Gemma3 models. Gemma3 models also introduced character-level naming errors and spatial mislocations that Qwen3-VL models did not produce. If a model cannot reliably extract map content, map access cannot produce the content-driven over-alignment bias we observe in Qwen3-VL.}

Figures~\ref{fig:scaling-qwen}, \ref{fig:scaling-gemma}, and \ref{fig:model-status} in Appendix~\ref{sec:app-model-details} show the full condition-level breakdown.
The status-level analysis (Figure~\ref{fig:model-status}) reveals that the aligned vs.\ non-aligned case trade-off observed for Qwen3-VL-8B generalises across models: map access consistently improves aligned-case F1 while degrading performance on pending and misunderstood REs, with the exception of Qwen3-VL-4B, where the overall conservative shift suppresses performance across all status categories.

\paragraph{Model size does not predict task performance.}
Within both families, the scaling relationship is non-monotonic: Qwen3-VL-2B outperforms 4B on both-maps (.566 vs.\ .411), and Gemma-3-4B outperforms 12B (.510 vs.\ .416).
Calibration data reveals why: Qwen3-VL-2B achieves the best calibration (ECE = .144) because it is the least confident (mean confidence .789), while Gemma-3-12B achieves the worst (ECE = .531) at the second highest confidence (.948).
Larger models become more confident without becoming more discerning. The bottleneck may be the tendency to commit to reference-related evidence-driven grounding establishment with excessive confidence.

%% file: sections/conclusion.tex
\section{Conclusion}\label{sec:conclusion}

\rev{We set out to test whether VLMs can judge \emph{interpretation matching} in information-asymmetric collaborative dialogue, using an evaluation methodology based on systematically controlled map-information and dialogue-context conditions applied to the HCRC MapTask.} The main result is not simply that maps help. Authentic maps improve overall performance, but they do so by pushing models toward \textsc{Yes}: landmark co-presence is treated as evidence of mutual understanding. Textual descriptions of the same map content reproduce the same bias, while non-informative visual inputs (blank, shuffled) reverse it into hyper-caution. The problem is therefore not multimodality in general, nor a specific visual-perceptual cue, but a tendency to read task-relevant information about \emph{potential} common ground as evidence that it has been \emph{established} through grounding.

Further analyses show where this bias is structured. Under map conditions, models become confidently biased toward aligned judgments, gain on already aligned cases while losing accuracy on pending and misunderstood ones, and grow more likely to predict alignment across repeated mentions.
Taken together, these patterns suggest that current VLMs are better at assessing potential referential overlap from map content than at tracking grounding as an interactional and incremental process. In MapTask terms, they can infer what the interlocutors \emph{could} be talking about, but they do not reliably distinguish this from what the interlocutors have actually established through grounding.

Our experiments are limited to an overhearer setting in a single domain, so an important next step is to test interactive settings where models can ask for clarification, express uncertainty, or revise their judgments over time. More broadly, extending the evaluation beyond MapTask and relating the observed heterogeneity to model properties via mechanistic interpretability analysis (e.g., attention flow analysis, \citealp{zhang2025cross}) \rev{may clarify whether the observed over-alignment is a general behavioural tendency in VLMs for collaborative dialogue, or a narrower failure mode tied to this task, setting, and/or model family.}

%% file: sections/limitations.tex
\section*{Limitations}

\paragraph{Generalization} Our primary evaluation relies on one single model (Qwen3-VL-8B-Instruct) for the detailed condition grid, with additional models tested only on baseline conditions.

\rev{The dataset derives from HCRC MapTask dialogues, a single corpus with specific properties that may shape the observed effects: a 72.1\%/27.9\% class imbalance toward aligned cases, MapTask-specific discrepancy types (missing, duplicated, or renamed landmarks), and a rare misunderstood category (239 of 13{,}077 instances). Suitable datasets for replication should provide perspectivist annotations recording both interlocutors' referent interpretations separately, not just task success or dialogue-level grounding labels, which limits the pool of available corpora. Generalisation to other information-asymmetric settings remains to be established.}

\paragraph{Task Design} The evaluation places the model in an overhearer position, characterising a judgment failure rather than an interaction failure.

We use greedy decoding to elicit a binary judgment, which may not reflect the model's full distributional beliefs about alignment. A softer evaluation protocol (e.g., allowing multi-choice selection or free-form responses) might reveal a more nuanced picture.

\paragraph{Textual Map Reconstruction} Our textual conditions (\emph{Text-landmark-names} and \emph{Text-discrepancy-detail}) convey landmark name lists and inter-map discrepancies, but they do not fully reconstruct the spatial layout of the maps. The behavioral gap between textual and visual conditions may therefore partly reflect this incomplete reconstruction rather than a genuine visual-channel effect.

%% file: sections/acknowledgement.tex
\section*{Acknowledgments}

We appreciate the helpful comments and suggestions from the anonymous reviewers.
This work is funded by the Dutch Research Council (NWO) through the AiNed Fellowship Grant NGF.1607.22.002, \textit{Dealing with Meaning Variation in NLP}.

%% file: sections/availibility.tex
\section*{Code and Data Availability}

We release our code and prompt templates to facilitate reproducibility\footnote{\url{https://github.com/chnln/seeing-is-not-sharing}}.

%% file: sections/appendix.tex
\appendix

\section{Dataset and Annotation}\label{sec:app-dataset}

Our experiments use the perspectivist annotation release of the HCRC MapTask corpus~\citep{li2025grounded}, derived from the original corpus of \citet{anderson1991hcrc}. The release contains 13{,}077 reference expressions (REs) annotated across 128 dialogues, paired with 16 distinct map pairs (\texttt{m0}--\texttt{m15}, each used in 8 dialogues). For each RE the annotation records the giver's and follower's interpretations separately, which is what makes the \textsc{Yes}/\textsc{No} interpretation-matching task studied here well-defined.

\paragraph{Landmark discrepancy types}
The 16 map pairs contain four kinds of landmark variation: \textbf{identical} landmarks appear at the same position under the same name on both maps; \textbf{lexical variants} appear at the same position but under different names (e.g., \emph{white water} on the giver's map vs.\ \emph{rapids} on the follower's; 10 such pairs are documented); \textbf{existence discrepancies} (landmarks appearing on only one of the two maps); and \textbf{multiplicity discrepancies} (landmarks appearing \emph{twice} on one map but only once on the other; 16 such landmarks are documented, one per map pair). Existence and multiplicity discrepancies are the structural source of most misalignment in this corpus; lexical variants are resolved by name unification during annotation so they do not on their own produce \emph{misunderstood} REs.

\paragraph{Landmark ID format}
Because the original MapTask corpus uses the same landmark ID for all instances of same-named landmarks, the annotation release introduces a unified ID scheme that disambiguates multiplicity landmarks and tracks map-side provenance. Each landmark ID has the format
\begin{center}
\texttt{<map\_id>\_<concept>\#<ordinal>@<side>},
\end{center}
where \texttt{<map\_id>} is \texttt{m0}--\texttt{m15}; \texttt{<concept>} is the original landmark name (e.g., \texttt{diamond\_mine}); \texttt{\#<ordinal>} is present only for \emph{multiplicity} landmarks that appear twice on one map, with \texttt{0} denoting the lower/bottom instance and \texttt{1} the upper/top; and \texttt{@<side>} is \texttt{g} for the giver's map or \texttt{f} for the follower's. Examples: \texttt{m9\_stony\_desert@g}, \texttt{m9\_site\_of\_plane\_crash\#0@g}, \texttt{m2\_stone\_creek\#1@f}.

\paragraph{Annotation cascade}
Each RE is annotated along a five-step cascade that models incremental resolution. Each step is evaluated only when the preceding conditions are met; when the cascade terminates early, all downstream attributes are set to \texttt{null} in the released data.

\begin{table}[h]
\centering
\footnotesize
\setlength{\tabcolsep}{4pt}
\begin{tabular}{@{}c l l c@{}}
\toprule
\textbf{Step} & \textbf{Attribute} & \textbf{Perspective} & \textbf{Req.} \\
\midrule
1 & \texttt{is\_quantificational} & Speaker    & \texttt{false} \\
2 & \texttt{is\_specified}        & Addressee  & \texttt{true} \\
3 & \texttt{is\_accommodated}     & Addressee  & \texttt{true} \\
4 & \texttt{is\_grounded}         & Addressee  & \texttt{true} \\
5 & \texttt{is\_imagined}         & Addressee  & --- \\
\bottomrule
\end{tabular}
\caption{The five-step annotation cascade used to guide the LLM and improve the annotation quality~\citep{li2025grounded}. The \textbf{Req.}\ column shows the value an attribute must take for the cascade to proceed; step~5 is terminal.}
\label{tab:app-cascade}
\end{table}

\paragraph{Understanding states}
Each RE is assigned one of three understanding states, derived post-hoc from the cascade attributes and the match between the giver's and the follower's interpretation fields (after lexical-variant unification). In the released dataset: \textbf{aligned} (both participants ground the RE to the same or equivalent landmark) accounts for 9{,}435 REs (72.1\%); \textbf{pending} (the RE is quantificational, unspecified, unaccommodated, or otherwise ungrounded) accounts for 3{,}403 REs (26.0\%); and \textbf{misunderstood} (both participants believe they agree but ground to different landmarks) accounts for 239 REs (1.8\%). The main-text analyses in \S\ref{sec:status-tradeoff} decompose model performance along these three states.

\section{Experimental Setup and Reproducibility}\label{sec:app-setup}

All inference runs use vLLM~\citep{kwon2023efficient} as the backend with greedy decoding (temperature $= 0.0$, \texttt{random\_seed} $= 42$), \texttt{max\_new\_tokens} $= 16$, and constrained output via vLLM's logit masking to the \textsc{Yes} / \textsc{No} token choice. For the calibration analyses reported in \S\ref{sec:calibration}, we additionally record the top-20 logprobs per generation step. All experiments run on a single NVIDIA A100 80\,GB GPU. For every Qwen3-VL model we explicitly disable the built-in thinking mode to ensure a fair comparison with other models. All the models are accessed via the Hugging Face Transformers library~\citep{wolf-etal-2020-transformers}.

\section{Prompt Template}\label{sec:app-prompt}

Figure~\ref{fig:prompt-template} shows the full prompt template used for all conditions.

\begin{figure*}[h]
\centering
\fbox{\parbox{0.95\textwidth}{%
{\small\sffamily\bfseries System Prompt}\\[4pt]
{\small\ttfamily
You are a dialogue analysis expert. You are overhearing two participants doing a MapTask-style route navigation task.\\[4pt]
MapTask Background:\\
- Each participant has their own map and they cannot see each other's map.\\
- One participant (the Giver) describes a route using named landmarks on their map.\\
- The other participant (the Follower) tries to follow the instructions on their own map.\\
- The two maps may differ (some landmarks may be missing, duplicated, or placed differently), so the two participants can end up with different personal interpretations even if the dialogue sounds smooth.\\[4pt]
Task:\\
You will be given a dialogue context in which ONE target reference expression is marked with <<...>>.
Decide whether the two participants interpret that reference expression as pointing to the SAME specific landmark (interpretations match).\\[4pt]
Guidelines:\\
- Use only the provided dialogue context and (if available) the provided map image(s).\\
- A match can be supported by clear evidence of successful grounding (e.g., consistent descriptions, confirmations, coherent subsequent navigation).\\
- If the expression indicates quantificational asking, or unspecified/unresolved grounding, treat it as NOT a match.\\[4pt]
Output:\\
Answer with exactly one word: Yes or No.\\
- Yes = interpretations match.\\
- No = interpretations do not match.\\
Do not output anything else.\\[4pt]
Information you can access in this instance:\\
- Dialogue text: \$\{text\_access\}\\
- Map information: \$\{map\_access\}
}%
\vspace{6pt}\hrule\vspace{6pt}
{\small\sffamily\bfseries User Prompt}\\[4pt]
{\small\ttfamily
Below is the specific information for this judgement.\\[4pt]
Target reference expression:\\
\$\{target\_ref\}\\[4pt]
Dialogue context (the target RE is wrapped in << >>):\\
\$\{context\}\\[4pt]
Maps: if map images are provided, they appear below.
}}}
\caption{Prompt template. Template variables (\texttt{\$\{...\}}) are filled per instance based on the text-access and map-access conditions. For example, under the \textbf{startT} text-access window and \textbf{both-maps} access level, \texttt{\$\{text\_access\}} is filled with ``\emph{You can read the dialogue from the beginning of the conversation through the end of the transaction that contains the target reference expression (i.e., including the subsequent lines in that transaction after the target line).}'' and \texttt{\$\{map\_access\}} is filled with ``\emph{You are shown both the Giver's and the Follower's map images.}''; the two map images are appended after the user prompt.}
\label{fig:prompt-template}
\end{figure*}

\section{Textual Map-Information Examples}\label{sec:app-textual-mapinfo}

This section shows the text that fills the \texttt{\$\{map\_access\}} slot of the prompt template (Figure~\ref{fig:prompt-template}) for the two textual map-information conditions introduced in \S\ref{sec:exp-task}. Both examples are drawn from the same instance --- dialogue \texttt{q1ec1} (map pair \texttt{m12}), target RE \emph{a caravan park} --- so the two variants can be compared directly.

\paragraph{\texttt{text-landmark-names}.}
Under this condition, the model is given only the list of landmark names on each participant's map. The system prompt's map-information line reads: \emph{``You are given the list of landmark names on each participant's map (see below in the dialogue context).''} The \texttt{\$\{map\_access\}} slot of the user prompt is filled with the block shown in Figure~\ref{fig:app-landmark-names-example}.

\begin{figure*}[h]
\centering
\fbox{\parbox{0.95\textwidth}{%
{\small\ttfamily
Map landmark information:\\
- Giver's map landmarks: start, caravan park, old mill, abandoned cottage, fenced meadow, fenced meadow, west lake, trig point, monument, nuclear test site, east lake, farmed land, finish\\
- Follower's map landmarks: start, caravan park, picket fence, mill wheel, forest, abandoned cottage, fenced meadow, west lake, monument, golf course, east lake, farmed land
}}}
\caption{Example filling of \texttt{\$\{map\_access\}} under the \texttt{text-landmark-names} condition for dialogue \texttt{q1ec1}.}
\label{fig:app-landmark-names-example}
\end{figure*}

\paragraph{\texttt{text-discrepancy-detail}.}
In addition to the per-map landmark lists, this condition supplies an explicit textual summary of how the two maps differ (per-side exclusives, multiplicity landmarks, and shared landmarks). The system prompt's map-information line becomes: \emph{``You are given the landmark names on each participant's map and a description of how the two maps differ (see below in the dialogue context).''} The \texttt{\$\{map\_access\}} slot is filled with the block shown in Figure~\ref{fig:app-discrepancy-detail-example}.

\begin{figure*}[h]
\centering
\fbox{\parbox{0.95\textwidth}{%
{\small\ttfamily
Map landmark information:\\
- Giver's map landmarks: start, caravan park, old mill, abandoned cottage, fenced meadow, fenced meadow, west lake, trig point, monument, nuclear test site, east lake, farmed land, finish\\
- Follower's map landmarks: start, caravan park, picket fence, mill wheel, forest, abandoned cottage, fenced meadow, west lake, monument, golf course, east lake, farmed land\\
Discrepancies between maps:\\
- Landmarks on Giver's map ONLY (not on Follower's): finish, nuclear test site, old mill, trig point\\
- Landmarks on Follower's map ONLY (not on Giver's): forest, golf course, mill wheel, picket fence\\
- Landmarks appearing multiple times: fenced meadow appears 2 times on Giver's map\\
- Shared landmarks (on both maps): abandoned cottage, caravan park, east lake, farmed land, fenced meadow, monument, start, west lake
}}}
\caption{Example filling of \texttt{\$\{map\_access\}} under the \texttt{text-discrepancy-detail} condition for dialogue \texttt{q1ec1}.}
\label{fig:app-discrepancy-detail-example}
\end{figure*}

\section{Map Reading Sanity Check}\label{sec:app-map-reading}

Prior to the main experiments, we assessed whether the five VLMs can reliably read the hand-drawn MapTask map images by prompting each model with two tasks on all 32 maps: (1)~list all landmark names visible on the map, and (2)~describe each landmark's spatial position. Both tasks use greedy decoding with free-form text output (no constrained decoding).

\begin{figure*}[t]
\centering
\includegraphics[width=\textwidth]{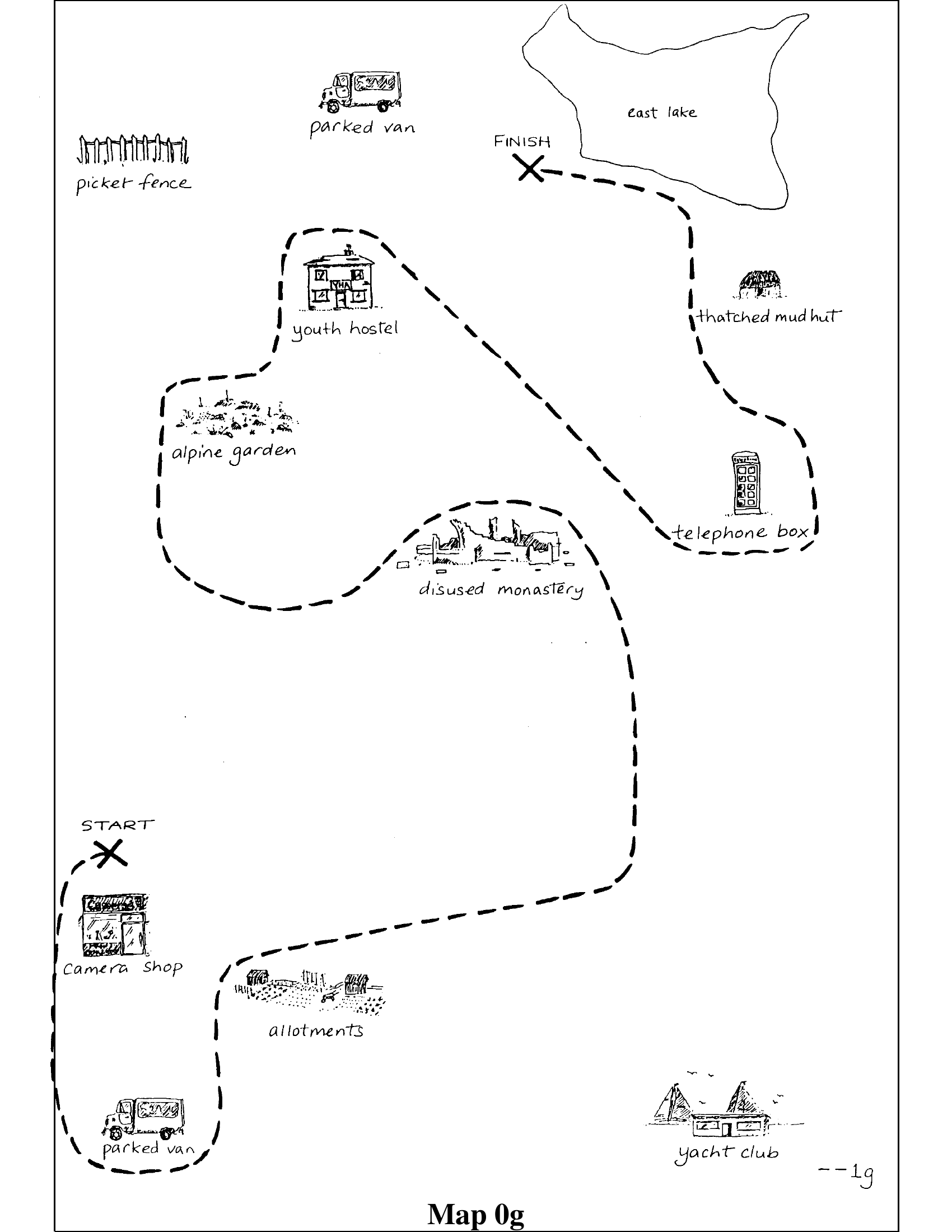}
\caption{Giver's map for map pair~0 (map0g), used for the spatial-description comparison in Table~\ref{tab:spatial-desc}.}
\label{fig:map0g}
\end{figure*}

\paragraph{Prompts.}
For task~1: \emph{``List all landmark names you can see on this map. Output only the landmark names, one per line. Do not add numbering, descriptions, or any other text.''}
For task~2: \emph{``For each landmark you can see on this map, describe its position on the map (e.g., top-left, upper-center, center, bottom-right). Format each line as: landmark name -- position. Do not add any other text.''}

\paragraph{Landmark listing.}
Table~\ref{tab:map-reading} reports recall, precision, and F1 for landmark-name identification across all 32 maps, evaluated against the gold landmark lists from the corpus metadata. Matching uses exact string comparison after lowercasing and whitespace normalisation.

\begin{table}[h]
\centering
\small
\begin{tabular}{l ccc}
\toprule
\textbf{Model} & \textbf{Recall} & \textbf{Precision} & \textbf{F1} \\
\midrule
Qwen3-VL-2B  & .845 & .910 & .876 \\
Qwen3-VL-4B  & .861 & .937 & .897 \\
Qwen3-VL-8B  & .812 & .956 & .878 \\
\midrule
Gemma-3-4B   & .766 & .866 & .813 \\
Gemma-3-12B  & .749 & .907 & .820 \\
\bottomrule
\end{tabular}
\caption{Landmark-name identification on all 32 MapTask map images. Recall = fraction of gold landmarks listed; Precision = fraction of model outputs that match a gold landmark.}
\label{tab:map-reading}
\end{table}

Qwen3-VL models achieve higher F1 (.876--.897) than Gemma3 models (.813--.820). Qwen3-VL-8B has the highest precision (.956), producing almost no spurious landmark names. Gemma3 models introduce character-level naming errors that Qwen3-VL avoids, such as ``picker fence'' instead of ``picket fence'' (both Gemma3 models) and ``pits of forest fire'' instead of ``site of forest fire'' (Gemma3-4B).

\begin{table}[t]
\centering
\footnotesize
\setlength{\tabcolsep}{2pt}
\begin{tabular}{l l l}
\toprule
\textbf{Landmark} & \textbf{Model} & \textbf{Position} \\
\midrule
\multirow{5}{*}{picket fence}
  & Qwen3-VL-2B & top-left \\
  & Qwen3-VL-4B & top-left \\
  & Qwen3-VL-8B & top-left \\
  & Gemma-3-4B  & upper-right$^\dag$ \\
  & Gemma-3-12B & top-left \\
\midrule
\multirow{5}{*}{east lake}
  & Qwen3-VL-2B & top-right \\
  & Qwen3-VL-4B & top-right \\
  & Qwen3-VL-8B & top-right \\
  & Gemma-3-4B  & bottom-right$^\dag$ \\
  & Gemma-3-12B & right-center$^\dag$ \\
\midrule
\multirow{5}{*}{\textsc{finish}}
  & Qwen3-VL-2B & top-right \\
  & Qwen3-VL-4B & top-right \\
  & Qwen3-VL-8B & top-right \\
  & Gemma-3-4B  & center$^\dag$ \\
  & Gemma-3-12B & right-center$^\dag$ \\
\midrule
\multirow{5}{*}{\textsc{start}}
  & Qwen3-VL-2B & --- \\
  & Qwen3-VL-4B & bottom-left \\
  & Qwen3-VL-8B & bottom-left \\
  & Gemma-3-4B  & center-left$^\dag$ \\
  & Gemma-3-12B & --- \\
\midrule
\multirow{5}{*}{camera shop}
  & Qwen3-VL-2B & bottom-left \\
  & Qwen3-VL-4B & bottom-left \\
  & Qwen3-VL-8B & bottom-left \\
  & Gemma-3-4B  & center-left$^\dag$ \\
  & Gemma-3-12B & top-left$^\dag$ \\
\midrule
\multirow{5}{*}{parked van (lower)}
  & Qwen3-VL-2B & bottom-left \\
  & Qwen3-VL-4B & bottom-left \\
  & Qwen3-VL-8B & bottom-left \\
  & Gemma-3-4B  & bottom-right$^\dag$ \\
  & Gemma-3-12B & bottom-left \\
\bottomrule
\end{tabular}
\caption{Spatial descriptions on map0g (Figure~\ref{fig:map0g}) for six landmarks where Gemma3 models make clear errors. Qwen3-VL models produce correct or near-correct positions on all 14 landmarks; only the six with Gemma3 errors are shown. ``---'' = not listed. $^\dag$ = position inconsistent with the map.}
\label{tab:spatial-desc}
\end{table}

\paragraph{Spatial description.}
We also prompted each model to describe landmark positions on all 32 maps. Figure~\ref{fig:map0g} shows the giver's map for map pair~0 (map0g), and Table~\ref{tab:spatial-desc} compares the outputs on this map for six landmarks where at least one Gemma3 model makes a clear spatial error. Qwen3-VL models produce correct or near-correct positions on all 14 landmarks; the six shown are selected to illustrate Gemma3's failure pattern. For example, Gemma-3-4B places east lake at bottom-right (it is at top-right), \textsc{start} and camera shop at center-left (both are at bottom-left), and picket fence at upper-right (it is at top-left). Gemma-3-12B similarly mislocates east lake and \textsc{finish}. Both Gemma3 models also output ``picker fence'' instead of ``picket fence''.

\section{Cross-Model Detailed Results}\label{sec:app-model-details}

\begin{figure*}[htbp]
\centering
\includegraphics[width=0.95\textwidth]{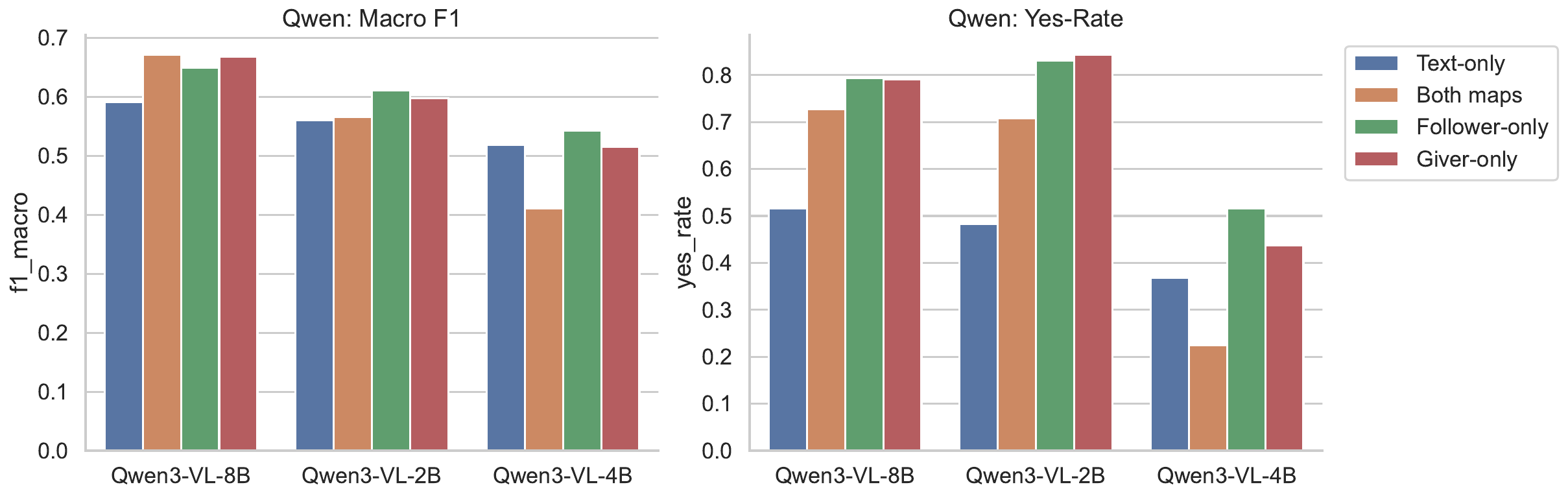}
\caption{Macro F1 and yes-rate across map-access conditions for Qwen3-VL models (8B, 2B, 4B) at startT.}
\label{fig:scaling-qwen}
\end{figure*}

\begin{figure*}[htbp]
\centering
\includegraphics[width=0.95\textwidth]{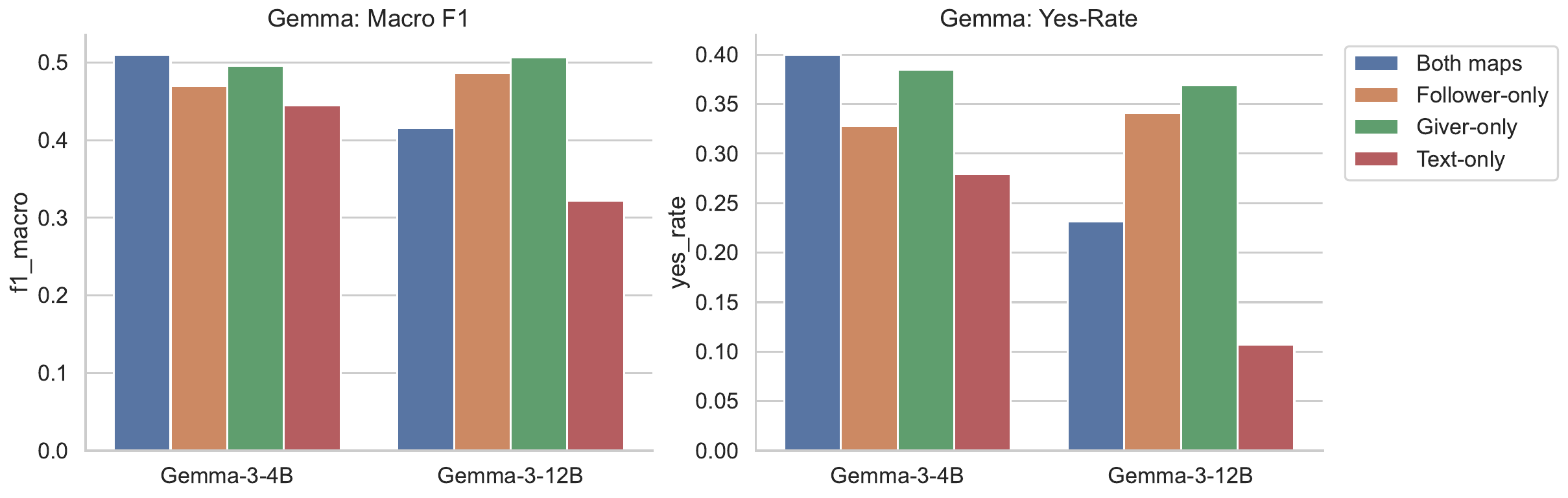}
\caption{Macro F1 and yes-rate across map-access conditions for Gemma-3 models (4B, 12B) at startT.}
\label{fig:scaling-gemma}
\end{figure*}

\begin{figure*}[htbp]
\centering
\includegraphics[width=0.95\textwidth]{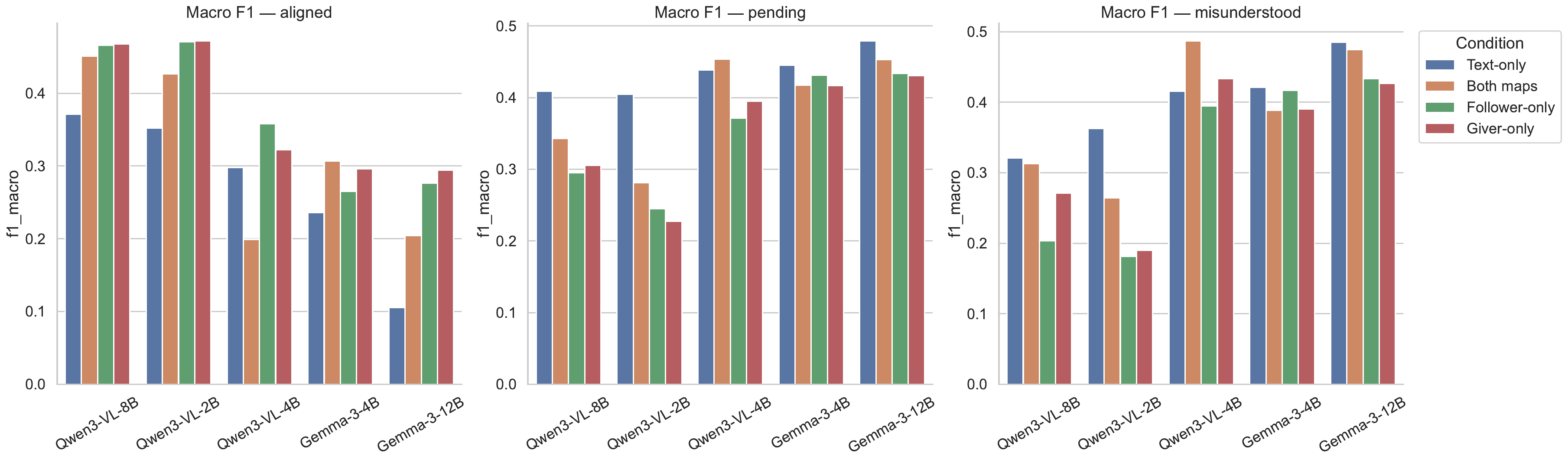}
\caption{Macro F1 by grounding status (aligned, pending, misunderstood) across all models and map-access conditions at startT.}
\label{fig:model-status}
\end{figure*}

Figures~\ref{fig:scaling-qwen}, \ref{fig:scaling-gemma} and \ref{fig:model-status} provide detailed breakdowns of the cross-model comparison in \S\ref{sec:model-comparison}.

%% file: custom.bib
@article{anderson1991hcrc,
  title={The {HCRC} {Map} {Task} corpus},
  author={Anderson, Anne H. and Bader, Miles and Bard, Ellen Gurman and Boyle, Elizabeth and Doherty, Gwyneth and Garrod, Simon and Isard, Stephen and Kowtko, Jacqueline and McAllister, Jan and Miller, Jim and Sotillo, Catherine and Thompson, Henry S. and Weinert, Regina},
  journal={Language and Speech},
  volume={34},
  number={4},
  pages={351--366},
  year={1991},
  publisher={SAGE Publications},
  doi={10.1177/002383099103400404},
  url={https://doi.org/10.1177/002383099103400404}
}

@inproceedings{li2025grounded,
  title={Grounded Misunderstandings in Asymmetric Dialogue: A Perspectivist Annotation Scheme for {MapTask}},
  author={Li, Nan and Gatt, Albert and Poesio, Massimo},
  booktitle={Proceedings of the Fifteenth Language Resources and Evaluation Conference (LREC 2026)},
  pages={4988--5001},
  year={2026},
  month=may,
  address={Palma, Mallorca, Spain},
  publisher={European Language Resources Association (ELRA)},
  doi={10.63317/59anbt78wyj7},
  url={https://lrec.elra.info/lrec2026-main-392}
}

@misc{bai2025qwen3vltechnicalreport,
  title={{Qwen3-VL} Technical Report},
  author={Shuai Bai and Yuxuan Cai and Ruizhe Chen and Keqin Chen and Xionghui Chen and Zesen Cheng and Lianghao Deng and Wei Ding and Chang Gao and Chunjiang Ge and Wenbin Ge and Zhifang Guo and Qidong Huang and Jie Huang and Fei Huang and Binyuan Hui and Shutong Jiang and Zhaohai Li and Mingsheng Li and Mei Li and Kaixin Li and Zicheng Lin and Junyang Lin and Xuejing Liu and Jiawei Liu and Chenglong Liu and Yang Liu and Dayiheng Liu and Shixuan Liu and Dunjie Lu and Ruilin Luo and Chenxu Lv and Rui Men and Lingchen Meng and Xuancheng Ren and Xingzhang Ren and Sibo Song and Yuchong Sun and Jun Tang and Jianhong Tu and Jianqiang Wan and Peng Wang and Pengfei Wang and Qiuyue Wang and Yuxuan Wang and Tianbao Xie and Yiheng Xu and Haiyang Xu and Jin Xu and Zhibo Yang and Mingkun Yang and Jianxin Yang and An Yang and Bowen Yu and Fei Zhang and Hang Zhang and Xi Zhang and Bo Zheng and Humen Zhong and Jingren Zhou and Fan Zhou and Jing Zhou and Yuanzhi Zhu and Ke Zhu},
  year={2025},
  eprint={2511.21631},
  archivePrefix={arXiv},
  primaryClass={cs.CV},
  url={https://arxiv.org/abs/2511.21631}, 
}

@inproceedings{kwon2023efficient,
  title={Efficient Memory Management for Large Language Model Serving with {PagedAttention}},
  author={Kwon, Woosuk and Li, Zhuohan and Zhuang, Siyuan and Sheng, Ying and Zheng, Lianmin and Yu, Cody Hao and Gonzalez, Joseph and Zhang, Hao and Stoica, Ion},
  booktitle={Proceedings of the 29th Symposium on Operating Systems Principles},
  pages={611--626},
  year={2023},
  month=oct,
  publisher={ACM},
  doi={10.1145/3600006.3613165},
  url={https://doi.org/10.1145/3600006.3613165}
}

@misc{kamath2025gemma,
  title={{Gemma} 3 Technical Report},
  author={{Gemma Team}},
  year={2025},
  eprint={2503.19786},
  archivePrefix={arXiv},
  primaryClass={cs.CL},
  url={https://arxiv.org/abs/2503.19786}
}

@inproceedings{naeini2015obtaining,
  title={Obtaining Well Calibrated Probabilities Using {Bayesian} Binning},
  author={Pakdaman Naeini, Mahdi and Cooper, Gregory and Hauskrecht, Milos},
  booktitle={Proceedings of the AAAI Conference on Artificial Intelligence},
  volume={29},
  number={1},
  year={2015},
  publisher={Association for the Advancement of Artificial Intelligence},
  doi={10.1609/aaai.v29i1.9602},
  url={https://ojs.aaai.org/index.php/AAAI/article/view/9602}
}

@inproceedings{guo2017calibration,
  title={On Calibration of Modern Neural Networks},
  author={Guo, Chuan and Pleiss, Geoff and Sun, Yu and Weinberger, Kilian Q.},
  booktitle={Proceedings of the 34th International Conference on Machine Learning},
  pages={1321--1330},
  editor={Precup, Doina and Teh, Yee Whye},
  volume={70},
  series={Proceedings of Machine Learning Research},
  year={2017},
  publisher={PMLR},
  url={https://proceedings.mlr.press/v70/guo17a.html}
}

@inproceedings{zhang2025cross,
  title={Cross-modal Information Flow in Multimodal Large Language Models},
  author={Zhang, Zhi and Yadav, Srishti and Han, Fengze and Shutova, Ekaterina},
  booktitle={2025 IEEE/CVF Conference on Computer Vision and Pattern Recognition (CVPR)},
  pages={19781--19791},
  year={2025},
  month=jun,
  publisher={IEEE},
  doi={10.1109/CVPR52734.2025.01842},
  url={https://doi.org/10.1109/CVPR52734.2025.01842}
}

@article{clark1986ReferringCollaborativeProcess,
  title = {Referring as a Collaborative Process},
  author = {Clark, Herbert H. and Wilkes-Gibbs, Deanna},
  year = {1986},
  journal = {Cognition},
  volume = {22},
  number = {1},
  pages = {1--39},
  publisher = {Elsevier},
  doi = {10.1016/0010-0277(86)90010-7},
  url = {https://doi.org/10.1016/0010-0277(86)90010-7},
  urldate = {2024-09-12},
  langid = {american}
}

@article{clark1989ContributingDiscourse,
  title = {Contributing to Discourse},
  author = {Clark, Herbert H. and Schaefer, Edward F.},
  year = {1989},
  journal = {Cognitive Science},
  shortjournal = {Cognitive Science},
  volume = {13},
  number = {2},
  pages = {259--294},
  issn = {0364-0213},
  doi = {10.1207/s15516709cog1302_7},
  url = {https://doi.org/10.1207/s15516709cog1302_7},
  urldate = {2025-10-20},
  langid = {american}
}

@incollection{clark1991GroundingCommunication,
  title = {Grounding in Communication.},
  booktitle = {Perspectives on Socially Shared Cognition.},
  author = {Clark, Herbert H. and Brennan, Susan E.},
  editor = {Resnick, Lauren B. and Levine, John M. and Teasley, Stephanie D.},
  year = {1991},
  pages = {127--149},
  publisher = {American Psychological Association},
  location = {Washington},
  doi = {10.1037/10096-006},
  url = {http://content.apa.org/books/10096-006},
  urldate = {2024-09-12},
  isbn = {978-1-55798-121-9},
  langid = {english}
}

@inproceedings{udagawa2019NaturalLanguageCorpus,
  title={A Natural Language Corpus of Common Grounding under Continuous and Partially-Observable Context},
  author={Udagawa, Takuma and Aizawa, Akiko},
  booktitle={Proceedings of the AAAI Conference on Artificial Intelligence},
  volume={33},
  number={01},
  pages={7120--7127},
  year={2019},
  publisher={Association for the Advancement of Artificial Intelligence},
  doi={10.1609/aaai.v33i01.33017120},
  url={https://doi.org/10.1609/aaai.v33i01.33017120}
}

@article{schober1989UnderstandingAddresseesOverhearers,
  title = {Understanding by Addressees and Overhearers},
  author = {Schober, Michael F and Clark, Herbert H},
  year = {1989},
  journal = {Cognitive Psychology},
  shortjournal = {Cognitive Psychology},
  volume = {21},
  number = {2},
  pages = {211--232},
  issn = {00100285},
  doi = {10.1016/0010-0285(89)90008-X},
  url = {https://linkinghub.elsevier.com/retrieve/pii/001002858990008X},
  urldate = {2024-09-03},
  langid = {english}
}

@inproceedings{haber2019photobook,
  title={The {P}hoto{B}ook Dataset: Building Common Ground through Visually-Grounded Dialogue},
  author={Haber, Janosch and Baumg{\"a}rtner, Tim and Takmaz, Ece and Gelderloos, Lieke and Bruni, Elia and Fern{\'a}ndez, Raquel},
  booktitle={Proceedings of the 57th Annual Meeting of the Association for Computational Linguistics},
  pages={1895--1910},
  year={2019},
  month=jul,
  address={Florence, Italy},
  publisher={Association for Computational Linguistics},
  doi={10.18653/v1/P19-1184},
  url={https://aclanthology.org/P19-1184/}
}

@inproceedings{wolf-etal-2020-transformers,
    title = "Transformers: State-of-the-Art Natural Language Processing",
    author = "Wolf, Thomas  and
      Debut, Lysandre  and
      Sanh, Victor  and
      Chaumond, Julien  and
      Delangue, Clement  and
      Moi, Anthony  and
      Cistac, Pierric  and
      Rault, Tim  and
      Louf, Remi  and
      Funtowicz, Morgan  and
      Davison, Joe  and
      Shleifer, Sam  and
      von Platen, Patrick  and
      Ma, Clara  and
      Jernite, Yacine  and
      Plu, Julien  and
      Xu, Canwen  and
      Le Scao, Teven  and
      Gugger, Sylvain  and
      Drame, Mariama  and
      Lhoest, Quentin  and
      Rush, Alexander",
    booktitle = "Proceedings of the 2020 Conference on Empirical Methods in Natural Language Processing: System Demonstrations",
    month = oct,
    year = "2020",
    address = "Online",
    publisher = "Association for Computational Linguistics",
    url = "https://aclanthology.org/2020.emnlp-demos.6/",
    doi = "10.18653/v1/2020.emnlp-demos.6",
    pages = "38--45"
}

@inproceedings{madureira-schlangen-2024-couldnt,
    title = "It Couldn{'}t Help but Overhear: On the Limits of Modelling Meta-Communicative Grounding Acts with Supervised Learning",
    author = "Madureira, Brielen  and
      Schlangen, David",
    editor = "Kawahara, Tatsuya  and
      Demberg, Vera  and
      Ultes, Stefan  and
      Inoue, Koji  and
      Mehri, Shikib  and
      Howcroft, David  and
      Komatani, Kazunori",
    booktitle = "Proceedings of the 25th Annual Meeting of the Special Interest Group on Discourse and Dialogue",
    month = sep,
    year = "2024",
    address = "Kyoto, Japan",
    publisher = "Association for Computational Linguistics",
    url = "https://aclanthology.org/2024.sigdial-1.13/",
    doi = "10.18653/v1/2024.sigdial-1.13",
    pages = "149--158"
}

@inproceedings{wang-etal-2025-lvlms,
    title = "{LVLM}s are Bad at Overhearing Human Referential Communication",
    author = "Wang, Zhengxiang  and
      Li, Weiling  and
      Kaliosis, Panagiotis  and
      Rambow, Owen  and
      Brennan, Susan",
    editor = "Christodoulopoulos, Christos  and
      Chakraborty, Tanmoy  and
      Rose, Carolyn  and
      Peng, Violet",
    booktitle = "Proceedings of the 2025 Conference on Empirical Methods in Natural Language Processing",
    month = nov,
    year = "2025",
    address = "Suzhou, China",
    publisher = "Association for Computational Linguistics",
    url = "https://aclanthology.org/2025.emnlp-main.849/",
    doi = "10.18653/v1/2025.emnlp-main.849",
    pages = "16758--16782",
    ISBN = "979-8-89176-332-6"
}

@misc{zeng2026lvlms,
  title={{LVLMs} and Humans Ground Differently in Referential Communication},
  author={Zeng, Peter and Li, Weiling and Paige, Amie J. and Wang, Zhengxiang and Kaliosis, Panagiotis and Samaras, Dimitris and Zelinsky, Gregory and Brennan, Susan E. and Rambow, Owen},
  year={2026},
  eprint={2601.19792},
  archivePrefix={arXiv},
  primaryClass={cs.CL},
  url={https://arxiv.org/abs/2601.19792}
}

@inproceedings{shaikh-etal-2025-navigating,
    title = "Navigating Rifts in Human-{LLM} Grounding: Study and Benchmark",
    author = "Shaikh, Omar  and
      Mozannar, Hussein  and
      Bansal, Gagan  and
      Fourney, Adam  and
      Horvitz, Eric",
    editor = "Che, Wanxiang  and
      Nabende, Joyce  and
      Shutova, Ekaterina  and
      Pilehvar, Mohammad Taher",
    booktitle = "Proceedings of the 63rd Annual Meeting of the Association for Computational Linguistics (Volume 1: Long Papers)",
    month = jul,
    year = "2025",
    address = "Vienna, Austria",
    publisher = "Association for Computational Linguistics",
    url = "https://aclanthology.org/2025.acl-long.1016/",
    doi = "10.18653/v1/2025.acl-long.1016",
    pages = "20832--20847",
    ISBN = "979-8-89176-251-0"
}

@inproceedings{chiyah-garcia-etal-2023-referring,
    title = "`{W}hat are you referring to?' {E}valuating the Ability of Multi-Modal Dialogue Models to Process Clarificational Exchanges",
    author = "Chiyah-Garcia, Javier  and
      Suglia, Alessandro  and
      Eshghi, Arash  and
      Hastie, Helen",
    editor = "Stoyanchev, Svetlana  and
      Joty, Shafiq  and
      Schlangen, David  and
      Dusek, Ondrej  and
      Kennington, Casey  and
      Alikhani, Malihe",
    booktitle = "Proceedings of the 24th Annual Meeting of the Special Interest Group on Discourse and Dialogue",
    month = sep,
    year = "2023",
    address = "Prague, Czechia",
    publisher = "Association for Computational Linguistics",
    url = "https://aclanthology.org/2023.sigdial-1.16/",
    doi = "10.18653/v1/2023.sigdial-1.16",
    pages = "175--182"
}
